\pdfoutput=1

\documentclass[11pt]{article}

\usepackage[review]{emnlp2021}

\usepackage{times}
\usepackage{latexsym}

\usepackage[T1]{fontenc}

\usepackage{booktabs}
\usepackage[utf8]{inputenc}

\usepackage{microtype}

\usepackage{times}
\usepackage{latexsym}
\usepackage{natbib}
\usepackage{amsmath}
\usepackage{xspace}
\usepackage{bbm}
\usepackage{graphicx}
\usepackage{color}
\usepackage{amsmath}
\usepackage{amssymb}
\usepackage{booktabs}
\usepackage{graphicx}
\usepackage{hyperref}
\usepackage{relsize}
\usepackage{latexsym}
\usepackage{paralist}
\usepackage{times}
\usepackage[printwatermark]{xwatermark}

\usepackage{url}
\usepackage{xcolor}
\usepackage{xspace}
\usepackage{diagbox}
\usepackage{multirow}
\usepackage{makecell}
\usepackage{bigdelim}
\usepackage{mathtools}

\usepackage{booktabs}
\usepackage{microtype}

\aclfinalcopy 

\newcommand\reddit{\textsc{Reddit}\xspace}
\newcommand\legal{\textsc{Legal}\xspace}
\newcommand\med{\textsc{Med}\xspace}
\newcommand\cs{\textsc{CS}\xspace}
\newcommand\realnews{\textsc{RealNews}\xspace}
\newcommand\webtext{\textsc{WebText}\xspace}
\newcommand\reviews{\textsc{Reviews}\xspace}

\newcommand\gptthree{\textsc{GPT-3}\xspace}

\newcommand\target{\textsc{Target}\xspace}

\newcommand\dapt{\textsc{DAPT}\xspace}
\newcommand\demixlayer{\textsc{DEMix} layer\xspace}
\newcommand\demixlayers{\textsc{DEMix} layers\xspace}
\newcommand\demix{\textsc{DEMix}\xspace}
\newcommand\plusexpert{\textsc{+Expert}\xspace}
\newcommand\minusexpert{\textsc{--Expert}\xspace}
\newcommand\minusdomain{\textsc{--Domain}\xspace}

\newcommand\demixdapt{\textsc{DEMix-DAPT}\xspace}
\newcommand\densedapt{\textsc{Dense-DAPT}\xspace}

\newcommand\dense{\textsc{Dense}\xspace}

\newcommand\domaintoken{\textsc{+Domain-Token}\xspace}

\newcommand\gutenberg{\textsc{Gutenberg}\xspace}
\newcommand\github{\textsc{Github}\xspace}
\newcommand\tweets{\textsc{Tweets}\xspace}
\newcommand\cord{\textsc{CORD-19}\xspace}
\newcommand\breakingnews{\textsc{Breaking News}\xspace}
\newcommand\yelp{\textsc{Yelp Reviews}\xspace}
\newcommand\aclpapers{\textsc{ACL Papers}\xspace}
\newcommand\contracts{\textsc{Contracts}\xspace}

\newcommand\oneb{\textsc{1B}\xspace}

\title{\demix Layers: Disentangling Domains for Modular Language Modeling}

\author{
    Suchin Gururangan$^{\dagger\diamondsuit}$ \quad
	Mike Lewis$^{\diamondsuit}$ \quad 
	\bf Ari Holtzman$^{\dagger}$ \quad 
	\bf Noah A. Smith$^{\dagger\spadesuit}$ \quad
	\bf Luke Zettlemoyer$^{\dagger\diamondsuit}$ \\
	$^\dagger$Paul G. Allen School of Computer Science \& Engineering, University of Washington \\
	$^\spadesuit$Allen Institute for AI \\
	$^\diamondsuit$Facebook AI Research \\
    Seattle, WA, USA \\
	{\tt sg01@cs.washington.edu }
}

\begin{document}
\maketitle
\begin{abstract}

We introduce a new domain expert mixture (\demix) layer that enables conditioning a language model (LM) on the domain of the input text.  A \demixlayer is a collection of expert feedforward networks, each specialized to a domain, that makes the LM \emph{modular}: experts can be mixed, added or removed after initial training. Extensive experiments with autoregressive transformer LMs (up to 1.3B parameters) show that \demixlayers reduce test-time perplexity, increase training efficiency, and enable rapid adaptation with little overhead. We show that mixing experts during inference, using a parameter-free weighted ensemble, allows the model to better generalize to heterogeneous or unseen domains. We also show that experts can be added to iteratively incorporate new domains without forgetting older ones, and that experts can be removed to restrict access to unwanted domains, without additional training. Overall, these results demonstrate benefits of explicitly conditioning on textual domains during language modeling.

\end{abstract}

\section{Introduction}

\begin{figure*}[t]
    \centering
    \includegraphics[scale=0.25]{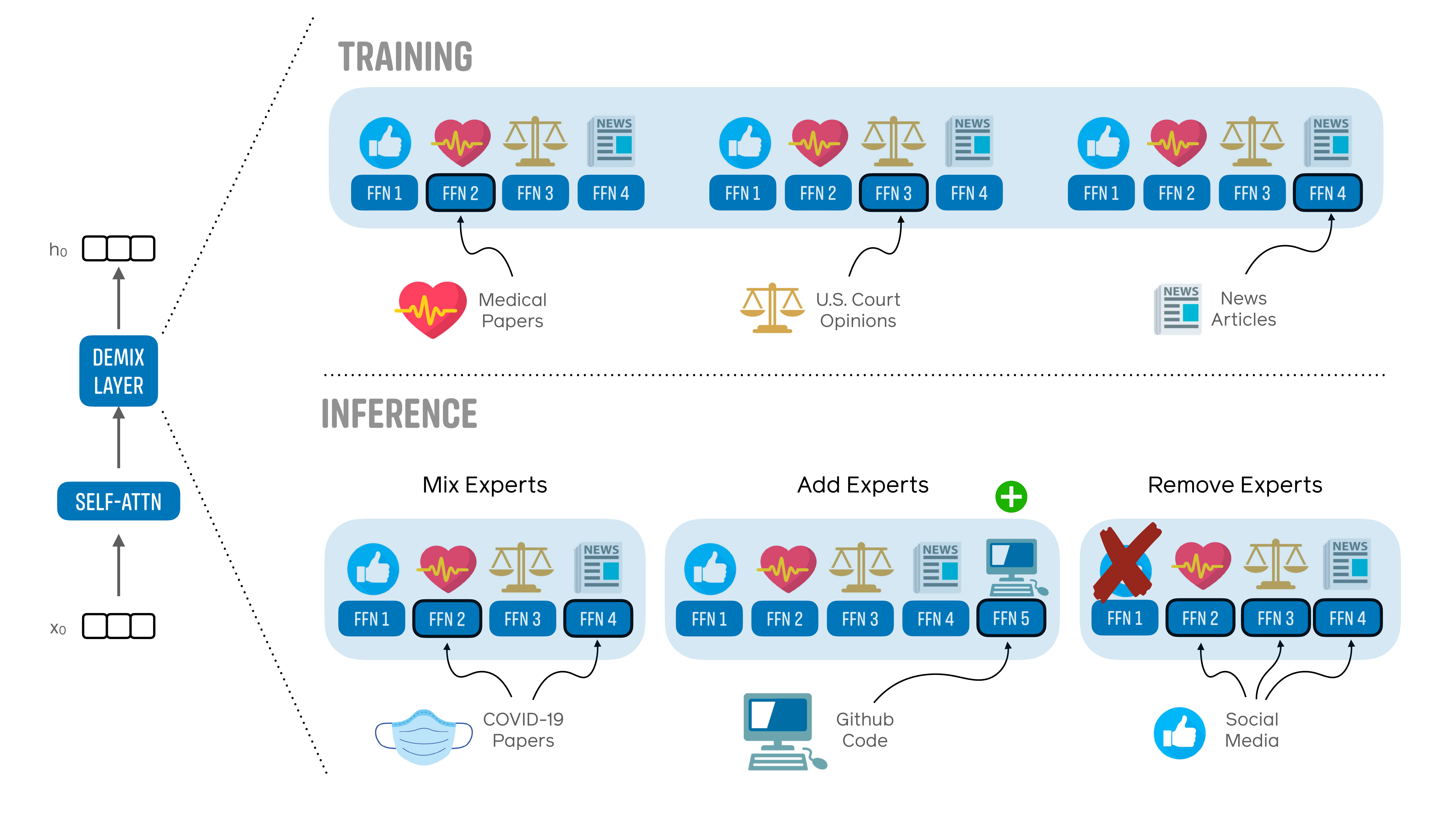}
    \caption{Illustration of a  \demixlayer in a single transformer block. During training, expert feedforward networks are conditionally activated based on the domain (here, document provenance) of the input sequence (i.e., scientific papers or court opinions). At inference time, the language model has new modular functions: domain experts can be mixed to handle heterogeneous domains, added to adapt to novel domains, or removed to ``forget'' unwanted domains. Image attribution: news icon from emojipedia.org; all other icons from istockphoto.com.}
    \label{fig:schwartz_plot}
\end{figure*}




Conventional language model (LM) training algorithms assume data homogeneity:  all parameters are updated to minimize the loss on all of the data. We refer to this approach as \emph{dense training}.  Yet human language is as varied as human experience, a fact researchers often refer to obliquely when they use the term \emph{domain} to describe distinct underlying subpopulations of the corpus. Dense training leaves variation in the data to be implicitly discovered \citep{aharoni-goldberg-2020-unsupervised}, assuming that models will be able to fit all domains equally well.



 While dense training is convenient, and densely trained LMs achieve impressive results \citep{brown2020language}, the approach has drawbacks with respect to generalization, efficiency, and flexibility. Even if training data is sourced from many domains, dense training can in practice emphasize subsets of the data  in proportion to their ease of access \citep{oren-etal-2019-distributionally, fan2020englishcentric}, limiting generalization to less prevalent domains. Updating all parameters of the network gets substantially more expensive as model size grows \citep{strubell-etal-2019-energy}, making fine-tuning or domain-adaptive pretraining (\dapt; \citealp{gururangan-etal-2020-dont}) harder to perform with smaller computational budgets. 
 It is also difficult to adapt to new domains without forgetting the original data \citep{McCloskey1989CatastrophicII, Aghajanyan2021BetterFB} or 
 restrict access to certain domains the LM has been exposed to during training (e.g., those that contain hate speech; \citealt{10.1145/3442188.3445922}), leading to risks of unwanted behavior \citep{gehman-etal-2020-realtoxicityprompts}.

To address these limitations of dense training, we argue that LMs should be designed with \emph{modularity}. We propose a modular LM that has components specialized to distinct domains in the training data, and can be customized at inference-time by mixing, adding, or removing these separated components as needed. This design principle emphasizes the ability to rapidly adapt the LM after training, a need that has been broadly advocated for language systems \citep{dinan2021anticipating, lazaridou2021pitfalls}.

We introduce modularity into an LM with a new domain expert (\demix) layer that explicitly conditions the LM on the domain of the input text (when it is known), or estimates the input domain during inference (when it is not known).  A \demix layer is a drop-in substitute for a feedforward layer in a transformer LM (e.g., \gptthree), creating a specialized version of the layer (or \emph{expert}) per domain (see Figure~\ref{fig:schwartz_plot}; \S\ref{sec:domain_parallel}).\footnote{This is an example of conditional computation \citep{fedus2021switch, lepikhin2020gshard, lewis2021base, roller2021hash}, which follow prior literature on mixture of experts \cite{hintonmixture, shazeer2017outrageously}.  Unlike dense training, conditional computation activates different parameters for different inputs. Instead of learning how to route data to experts, the \demix layer routing mechanism follows from a natural, observable segmentation of the data.} We find that replacing every feedforward layer in the transformer with a \demixlayer offers new affordances for modularity, addressing the challenges above, while improving performance in both training domains and novel test-time domains.

Although the concept of a domain lacks a rigorous definition in NLP, we use coarse provenance categories (e.g., whether a document is a medical research paper or a Reddit post) as a conditioning variable when training an LM with \demixlayers (\S\ref{sec:dataset}). Training on data from eight different domains, we find that \demixlayers consistently improve in-domain performance (\S\ref{sec:basic_experiment}). However, because these categories may not be an optimal segmentation of the training data, or may lack coverage of test-time domains, naively selecting a single domain expert at test time can hurt generalization. Instead, we introduce a parameter-free probabilistic approach to dynamically estimate a \emph{weighted mixture} of domains during inference (\S\ref{sec:mixing_experts}).  Mixing experts  improves \demix performance not only on \emph{novel} test-time domains, but also on test data from the \emph{training} domains, which may themselves be heterogeneous. Our results suggest that introducing modularity into an LM need not come at a cost to generalization performance.

Because \demix forces experts to specialize to domains, the overall model can be (partially) disentangled after training.  Beyond mixing, we can 
add (\S\ref{sec:adding_experts}) or remove (\S\ref{sec:removing_experts}) domain experts, resulting in predictable changes in model behavior at inference time:  adding experts allows for model adaptation without updating all parameters (hence avoiding forgetting), and removing experts allows for simulating the removal of training domains without additional training. Overall, \demixlayers demonstrate benefits of explicitly conditioning on textual domains during language modeling, and our results suggest that these benefits persist at scale. Our code is publicly available.\footnote{\url{http://github.com/kernelmachine/demix}}

\begin{table*}[t]
\centering
\small
\begin{tabular}{lllr}
\toprule
& \bf Domain & \bf Corpus &  \bf \# Train (Eval.) Tokens  \\
\midrule 


\parbox[t]{3pt}{\multirow{8}{*}{\rotatebox[origin=c]{90}{\bf {\textsc{Training}}}}} & \oneb 
& 30M NewsWire sentences \citep{chelba2014billion} 
& 700M (10M) \\
& \cs 
& 1.89M full-text CS papers from S2ORC \citep{lo-wang-2020-s2orc} 
& 4.5B  (10M)\\
& \legal
& 2.22M U.S. court opinions, 1658 to 2018 \citep{caselaw2018} 
& 10.5B (10M) \\
 & \med      
& 3.2M full-text medical papers from  S2ORC \citep{lo-wang-2020-s2orc}  
& 9.5B (10M)\\

& \textsc{WebText}$^\dagger$
& 8M Web documents \citep{Gokaslan2019OpenWeb} 
& 6.5B  (10M)\\

& \textsc{RealNews}$^\dagger$
& 35M articles from \realnews  \cite{Zellers2019DefendingAN}
& 15B (10M) \\
& \reddit
& Reddit comments from pushshift.io \citep{baumgartner2020pushshift} 
& 25B  (10M)\\

& \textsc{Reviews}$^\dagger$
& 30M Amazon product reviews \citep{ni-etal-2019-justifying} 
& 2.1B (10M) \\

\midrule
& & 
& \bf{Total} \hspace{2mm}  73.8B (80M) \\
\midrule
\end{tabular}

\begin{tabular}{lllr}
& \bf Domain & \bf  Corpus & \bf \# Train (Eval.) Tokens \\
\midrule

 \parbox[t]{3pt}{\multirow{8}{*}{\rotatebox[origin=c]{90}{\bf \textsc{Novel}}}} & \aclpapers & 1.5K NLP papers from ACL \citep{Dasigi2021ADO} & 1M (1M)  \\
 & \textsc{Breaking News}$^\dagger$ 
& 20K latest articles from 400 English news sites \citep{baly:2018:EMNLP2018}  & 11M (1M) \\

& \textsc{Contracts}$^\dagger$ & 500 commercial legal contracts \citep{hendrycks2021cuad} & 1.5M (1M) \\
& \cord 
& 400K excerpts from COVID-19 research papers \citep{wang2020cord19} &  60M (10M) \\
& \textsc{Github}
& 230K public Github repository contents  \citep{githubcode} & 200M (10M) \\
& \gutenberg      
& 3.2M copyright-expired books \citep{gutenberg} & 3B (10M) \\

& \textsc{Tweets}$^\dagger$ 
& 1M English tweets from 2013-2018 & 8M (1M)
\\

&  \textsc{Yelp Reviews}$^\dagger$ 
& 6M Yelp restaurant reviews \citep{yelpreviews} & 600M (10M)\\

\bottomrule
\end{tabular}

\caption{Domains that make up our multi-domain training corpus, including the size of our training and evaluation (i.e. validation and test) data, in whitespace-separated tokens. $\dagger$ indicates datasets that we (partially) anonymize (\S\ref{sec:dataset}). \reddit was extracted and obtained by a third party and made available on \url{pushshift.io}, and was anonymized by \citet{xu2020recipes}; we use their version. See Appendix \S\ref{sec:domain_collection} for more details on how these data were collected.}
\label{tab:datasets}
\end{table*}

\section{Multi-Domain Corpus}
\label{sec:dataset}

We center this study around a large, multi-domain corpus we constructed with explicit provenance metadata (Table \ref{tab:datasets}).  While other multi-domain corpora \citep{Koh2021-kr, gao2020pile} cover many more domains and tasks, the corpus we introduce contains substantial metadata-tagged text for language modeling, as well as datasets with friendly licensing to support reproducibility. 

\subsection{Document Provenance as a Domain Label}

While a growing body of work has attempted to address the structure and composition of language domains \citep{Eisenstein_2014,plank2016nonstandard,aharoni-goldberg-2020-unsupervised, gururangan-etal-2020-dont}, fundamentally what a domain is remains a matter of debate. In this work, we focus on the \emph{provenance} of a document, operationalized coarsely by the dataset we used to access it, which approximates a social process that produced it.  Defining domains this way is easy and intuitive, conveys a great deal about the variation in a document's language, and aligns with common practice in NLP research. However, other accounts of variation in language \citep[e.g.,][]{lucy2021characterizing}, and richer notions of relationships among domains  \citep[e.g., hierarchies;][]{gururangan-etal-2020-dont}, may be studied in future work.

\subsection{Corpus Description}

The multi-domain corpus we use in this study consists of two parts. The first is a collection of \textbf{training} domains: text from eight domains of largely English text, listed at the top of Table \ref{tab:datasets}, each of which vary in complexity and coverage and has been the subject of study in NLP.\footnote{The metadata for each document includes at least its provenance, and in some cases more information (e.g., URLs, publication venue, or legal jurisdiction).  Future work might explore more fine-grained notions of domain.}

The second part is a collection of \textbf{novel} domains: text from eight domains also of largely English text, listed at the bottom of Table \ref{tab:datasets},  which may or may not align with the training domains. The novel domains allow us to measure how models generalize to a more challenging data distribution shift, where domain boundaries may be less clear.

See Appendix \S\ref{sec:domain_collection} for more details on how these data were collected. To support future work with the data, we also release a standard API to download and preprocess it into a format compatible with Fairseq \citep{ott-etal-2019-fairseq}.\footnote{\url{https://github.com/kernelmachine/demix-data}}  We replace user identifiable information (e.g., email addresses, user handles, social security numbers, credit card numbers, phone numbers) with dummy tokens.\footnote{While it is difficult to anonymize data perfectly, especially at scale, we use a suite of regexes to identify commonly occurring identifiable information on the Internet. See Appendix \S\ref{sec:anonymization} for more details.}

\section{\demix Layer}
\label{sec:domain_parallel}


\subsection{Background: Mixture-of-Experts Transformers}

The transformer architecture is comprised of interleaved multi-head self-attention, layer-norms, and feedforward networks \citep{vaswani2017attention}.  Each of these layers produces a vector representation for each of the input tokens. Our focus is on the feedforward component:
\begin{align}
    \mathbf{h}_{t,\ell} = \mathrm{FFN}(\mathbf{h}_{t, \ell-1}),
\end{align}

\noindent
where $\mathbf{h}_{t,\ell}$ is the vector for the $t$th token produced by layer $\ell$.

\citet{shazeer2017outrageously} propose a formulation of one or more feedforward layers as an ensemble of $n$ experts $\mathrm{FFN}_1,\ldots,\mathrm{FFN}_n$, assigned weights respectively by functions $g_1,\ldots,g_n$:

\begin{align}
    \mathrm{FFN}(\mathbf{h}_{t,\ell-1}) &= \sum_{j=1}^n g_j(\mathbf{h}_{t,\ell-1}) \cdot \mathrm{FFN}_j(\mathbf{h}_{t,\ell-1})
\end{align}

The $g$ function routes tokens to different experts, usually each a separate instance of the original feedforward network.  If $g$ routes to a single expert, then the computational cost (in floating-point operations; FLOPs) will be same as the original feedforward network, even though it has slightly more than $n$ times as many parameters.

\subsection{\demix Routing}
\label{sec:domain_parallel_architecture}

Previous approaches \emph{learn} the weighting functions $g$ at a token-level, and either assign at most one \citep{fedus2021switch} or two \citep{lepikhin2020gshard} experts per token. This necessitates load balancing and other techniques to encourage the model to use all experts instead of relying on just a few \citep{fedus2021switch, lewis2021base}. 

We instead use domain metadata provided with training documents to route data to experts at the \emph{document} (i.e., sequence) level. During training, every token in the same sequence is assigned to the same expert based on the domain label.

Let $\mathcal{D}$ denote the set of domain labels (i.e., the eight labels in Table~\ref{tab:datasets}).  If we index the experts by $\mathcal{D}$ and $d \in \mathcal{D}$ is the domain label for the current training instance, then
\begin{align}
    g_j(\mathbf{h}_{t,\ell}) &= \left\{ \begin{array}{ll} 1 & \mbox{if $j=d$} \\
    0 & \mbox{otherwise} \end{array}\right. \label{eq:hard}
\end{align}

While we assume that each \emph{training} document is associated with a single domain label, we relax this requirement at inference time (\S\ref{sec:mixing_experts}), which improves model performance in mixed and unknown domain scenarios. 




\subsection{\demix Architecture}
\label{sec:demix_architecture}

Our design results in one expert in a \demixlayer per domain (i.e., eight experts for eight training domains in our multi-domain corpus). 

We replace \emph{every} feedforward layer in the transformer with a \demixlayer, in contrast to previous work  \citep{fedus2021switch,lepikhin2020gshard} that interleaves shared and expert layers.  Preliminary experiments showed that interleaving led to worse in-domain performance with \demix layers. We hypothesize that shared layers may serve as a bottleneck to find shared features between domains, and may impact performance adversely when training domains are highly different from one another.\footnote{Indeed, preliminary experiments suggest that interleaving expert layers causes large performance hits in the most distinct domains, i.e., those with lower vocabulary overlap with other domains in the corpus. } Future work might perform careful comparisons of different architectural choices. 

In this study, each expert $\mathrm{FFN}_j$ is a two-layer MLP with the same dimensions as the original $\mathrm{FFN}$ layer of the transformer. As with other conditional computation models \citep{fedus2021switch, lepikhin2020gshard}, this means that the effective number of parameters in the overall \demix LM increases (Table \ref{tab:exp_setup}). While this incurs memory costs, the computational budget we consider in this study centers around runtime costs. \demix layers decrease the runtime costs of training the LM. 

\subsection{\demix Training}
\label{sec:demix_training}

\demixlayers increase the total parameters of the LM while also reducing GPU latency costs during training, effectively reducing runtime costs of training the LM.

\dense training (also referred to as \emph{data-parallel}) is usually implemented by copying model parameters to every GPU, feeding a different mini-batch of shuffled data to each GPU, computing a stochastic gradient for each mini-batch, and updating all parameters synchronously with the average stochastic gradient from across all GPUs.

To train an LM with \demix layers, we instead partition the GPUs among the domains, so that each GPU is assigned a single domain (along with its corresponding expert). During training, we fill a mini-batch with $k$ sequences, where each sequence represents data from a particular domain, and we send each mini-batch to its dedicated domain expert. We use larger batch sizes by performing data-parallel training between expert parameters on GPUs assigned to the same domain; we assign $n$/8 GPUs to each domain (Table \ref{tab:exp_setup}). To reduce overfitting, we ensure that each of these $n$/8 GPUs is assigned to different shards of their domain's training data.


We compare the training efficiency of \dense and \demix  models up to 1.3B parameters per GPU in Table \ref{tab:exp_setup}. Compared to \dense LMs, \demixlayers achieve the same or slightly higher throughput (measured in TFLOPs/GPU) for the same total FLOPs per update, despite adding significantly more parameters.  

\demix achieves higher throughput because we only synchronize expert parameters allocated to the same domain.\footnote{Shared parameters are synchronized across all GPUs.} As we increase model size, this results in a reduction of latency costs between GPUs, and hence, faster training; instead of synchronizing parameters over $n$ GPUs, we perform eight synchronizations over $n$/8 GPUs.\footnote{While this technique reduces latency costs, the bandwidth costs are the same between \demix and \dense models.}



In this work, we assume that there is sufficient data for each training domain that each expert can be exposed to the same amount of data, and load balancing between experts is not necessary. Future work may consider how varying the amount of data per domain influences absolute and relative performance across domains, especially in the long tail of rare domains. 

While the total number of parameters of \demix LMs are substantially larger than their \dense counterparts, since the practical training costs are essentially the same, we compare baselines in all subsequent experiments based on parameters \emph{per GPU}, as we do in Table \ref{tab:exp_setup}. 


\section{In-Domain Performance}
\label{sec:basic_experiment}














\begin{table}[t]\small
\centering
\begin{tabular}{lrcccc}
\toprule
& & \multicolumn{4}{c}{\bf Parameters per GPU}\\
& & 125M & 350M & 760M & 1.3B \\
\cmidrule{3-6}
\parbox[t]{0.7pt}{\multirow{4}{*}{\rotatebox[origin=c]{90}{\bf {\textsc{\dense}}}}} & \bf GPUs  & 32 & 64 & 128 & 128\\
& \bf Total Experts & 0 & 0 & 0 & 0 \\
& \bf GPUs/expert & 0 & 0 & 0 & 0 \\
& \bf Total params & 125M & 350M & 760M & 1.3B \\
& \bf TFLOPs/update & 556 &  3279 & 13,637 & 23,250 \\
& \bf TFLOPs/GPU  & \bf{31} & \bf 37 & 45 & 51 \\
\cmidrule{3-6 }
\parbox[t]{0.7pt}{\multirow{4}{*}{\rotatebox[origin=c]{90}{\bf {\textsc{\demix}}}}} & \bf GPUs  & 32 & 64 & 128 & 128\\
& \bf Total Experts & 8 & 8 & 8 & 8 \\
& \bf GPUs/expert & 4 & 8 & 16 & 16 \\
& \bf Total params & 512M & 1.8B & 3.8B & 7.0B \\
& \bf TFLOPs/update & 556 & 3279 & 13,637 & 23,250 \\
& \bf TFLOPs/GPU & \bf 31 & \bf 37 & \bf 48 & \bf 55 \\




\bottomrule
\end{tabular}

\caption{Our specifications for training \dense and \demix LMs. All models are trained for about 48 hours on V100 GPUs. \demixlayers increase the total parameters of the LM while maintaining (or increasing) throughput, measured in TFLOPs/GPU. We use the formula described in \citet{narayanan2021efficient} to calculate these metrics. See Appendix \S\ref{sec:tflops} for more details.}
\label{tab:exp_setup}
\end{table}

The first set of experiments in this study considers the impact of replacing the conventional feedforward layers in a transformer LM with \demixlayers. We run all experiments in this section with the training domains (Table \ref{tab:datasets}).
\subsection{Experimental Setup}
\label{sec:basic_experiment_setup}

\paragraph{Architecture and Input}  The model architecture is a randomly-initialized LM with the \gptthree \citep{brown2020language} architecture implemented in Fairseq \citep{ott-etal-2019-fairseq}. We experiment with multiple architectures (i.e., those of \gptthree small, medium, large, and XL), at a maximum size of about 1.3B parameters per GPU. We use the GPT-2 \citep{radford2019language} vocabulary of 50,264 BPE types, and train with 1,024-token  sequences, with cross-document boundaries. Each document has a beginning-of-sentence token prepended to it.



\paragraph{Hyperparameters} We set the total number of training steps based on this allocated runtime, set 8\% of these steps to be warm-up, and use the Adam optimizer \citep{kingma2017adam} with a polynomial learning rate decay. Learning rates are tuned for each model separately over \{0.0001, 0.0003, 0.0005\}, taking the fastest learning rate that avoids divergence. Each worker processes two sequences of length 1,024, and gradients are accumulated over 8 updates. We clip gradients if their $L_2$ norm exceeds 0.1.  See Appendix \S\ref{sec:hyperparameters} for more details. These settings are inspired by \citet{lewis2021base}.

\paragraph{Computational Budget} We follow previous work in using runtime as the primary computational budget, which provides a better comparison of the practical costs of training conditional compute  and dense models \citep{lewis2021base}. We assume a fixed budget of about 48 hours on NVIDIA V100 32GB GPUs. We display the number of GPUs used for each model size in Table \ref{tab:exp_setup}; we chose these GPU budgets because larger models require more compute to train properly \citep{lewis2021base, kaplan2020scaling}, and found these GPU budgets to result in stable training for each model size given mostly fixed hyperparameters.

\paragraph{Evaluation} We report test-set perplexities after about 48 hours of training. In all tables, we report each result with respect to a set number of parameters per GPU, as in Table \ref{tab:exp_setup}. As mentioned in \S\ref{sec:demix_training}, \demix LM will have a larger effective size than the \dense LM at the same increased throughput.














\begin{table}[t]\small
\centering
\begin{tabular}{rcccc}
\toprule
& \multicolumn{4}{c}{\bf Parameters per GPU}\\
& 125M & 350M & 760M & 1.3B \\
\cmidrule{2-5}
\bf \dense  & 20.6 & 16.5 & 14.5 & 13.8\\
\bf \dense (Balanced)  & 19.9 & 15.8 & 14.3 & 13.6\\
\bf \domaintoken & 19.2 & 15.9 & 14.3 & \bf 13.4 \\
\bf \demix (naive) & 18.4 & 15.5 & 14.2 & 13.8\\
\bf \demix (cached; \S\ref{sec:improving_indomain}) & \bf 17.8 & \bf 14.7 & \bf 13.9 & \bf 13.4 \\



\bottomrule
\end{tabular}

\caption{Average of in-domain test-set perplexity. We discuss the last row in \S\ref{sec:improving_indomain}. } 
\label{tab:base_results}
\end{table}

\subsection{Compared Models}

\paragraph{\dense} The first baseline is a \dense model that treats the data as homogeneous, i.e., it shares all parameters across all domains. Under this setup, the language model parameters are copied across all GPUs, and gradients computed during training are all-reduced across every GPU. There is no explicit conditioning on domain.

\paragraph{\dense (Balanced)} Under this setting, we train densely but ensure that the model is exposed to an equal amount of data from each domain. While there is still no explicit conditioning on domain, the gradient updates that the model makes during training are an average of those computed across all domains represented in a batch.

\paragraph{\domaintoken} This model is trained identically to  \dense (Balanced), but we prepend a token indicating the sequence's domain to every sequence block (during training and test time). A variant of this domain token is explored in some previous studies \citep{Zellers2019DefendingAN, keskar2019ctrl}. 
This baseline provides domain information to the language model in the form of input supervision. We ignore the domain token when computing perplexity during evaluation.

\paragraph{\demix (naive)} We replace every feedforward layer in the transformer with a \demixlayer, as detailed in \S\ref{sec:domain_parallel}. Under this  setting, the domain of the test data is known and revealed to the model (e.g., the \cs expert is used for \cs test data), which we refer to as \emph{naive}. We also ensure that the model is exposed to an equal amount of data from each domain.

\begin{table}[t]\small
\centering
\begin{tabular}{rccc}
\toprule
&  \multicolumn{3}{c}{ \bf 1.3B parameters per GPU} \\
\cmidrule{2-4}

\textbf{Domain} & \bf \dense  & \bf \demix & \bf \demix \\ 
& & \bf (naive) & \bf (cached prior; \S\ref{sec:improving_indomain}) \\ 
\midrule

\oneb

& 11.8 & 11.5   & \bf 11.3       \\


\cs 

& 13.5 & 12.2 & \textbf{12.1} \\

\legal 
& 6.8  & \textbf{6.7} & \textbf{6.7} \\

\med  
& 9.5        & 9.2  & \textbf{9.1}  \\

\webtext 
& \bf 13.8 & 14.6  &  14.3      \\

\realnews 
& \bf 12.5 & 13.3   & 13.1       \\

\reddit 
& 28.4 & 30.6 &  \bf 28.1 \\

\reviews 
& 14.0 & 12.6 & \bf 12.5 \\

\midrule
\bf Average & 13.8 &  13.8 & \bf 13.4 \\

\bottomrule
\end{tabular}
\caption{Test-set perplexity by domain, for an LM with 1.3B parameters per GPU. We discuss the last column in \S\ref{sec:improving_indomain}. 
}
\label{tab:breakdown_results}
\end{table}

\subsection{Results}

Table \ref{tab:base_results} shows test-set perplexities, averaged across the eight training domains. First, we observe that domain balancing is consistently helpful for \dense training.  We find that balancing is especially important in cases in which there is an imbalance of domain prevalence, confirming similar observations from previous studies \citep{arivazhagan2019massively}.\footnote{Balancing improves performance on most domains, but hurts performance relative to a \dense baseline on the \reddit domain (Appendix \S\ref{sec:per_domain_results}). In the multi-domain corpus, there is far more \reddit text than anything else; see Table \ref{tab:datasets}.}

Next, we observe that the benefits of additional domain information (i.e, domain tokens or \demixlayers) are clearest for the smallest model; for larger models, the benefits are smaller but consistent. This result suggests that domain-specific information enables the model to better specialize to different domains in its training data. However, as the model size grows, the \dense baseline becomes increasingly better at fitting the training domains, catching up to models with additional domain information, in the average case.

\subsection{Domain Hetereogeneity}

A more complete view of the experiments with the largest model is shown in Table \ref{tab:breakdown_results}. We see that even at scale, most training domains benefit from \demixlayers in a naive setting (where the domain label is revealed at test time), but some do not; \webtext, \realnews, and \reddit fare worse than the \dense baseline. We believe that this variation can be explained by heterogeneity within domains and varying degrees of similarity between them. \dense training may be advantageous for domains that have a higher degree of overlap with other domains in the corpus (and therefore, benefit from parameter sharing).


\begin{figure}[t]
    \centering
    \hspace*{-0.5cm}
    \includegraphics[scale=0.55]{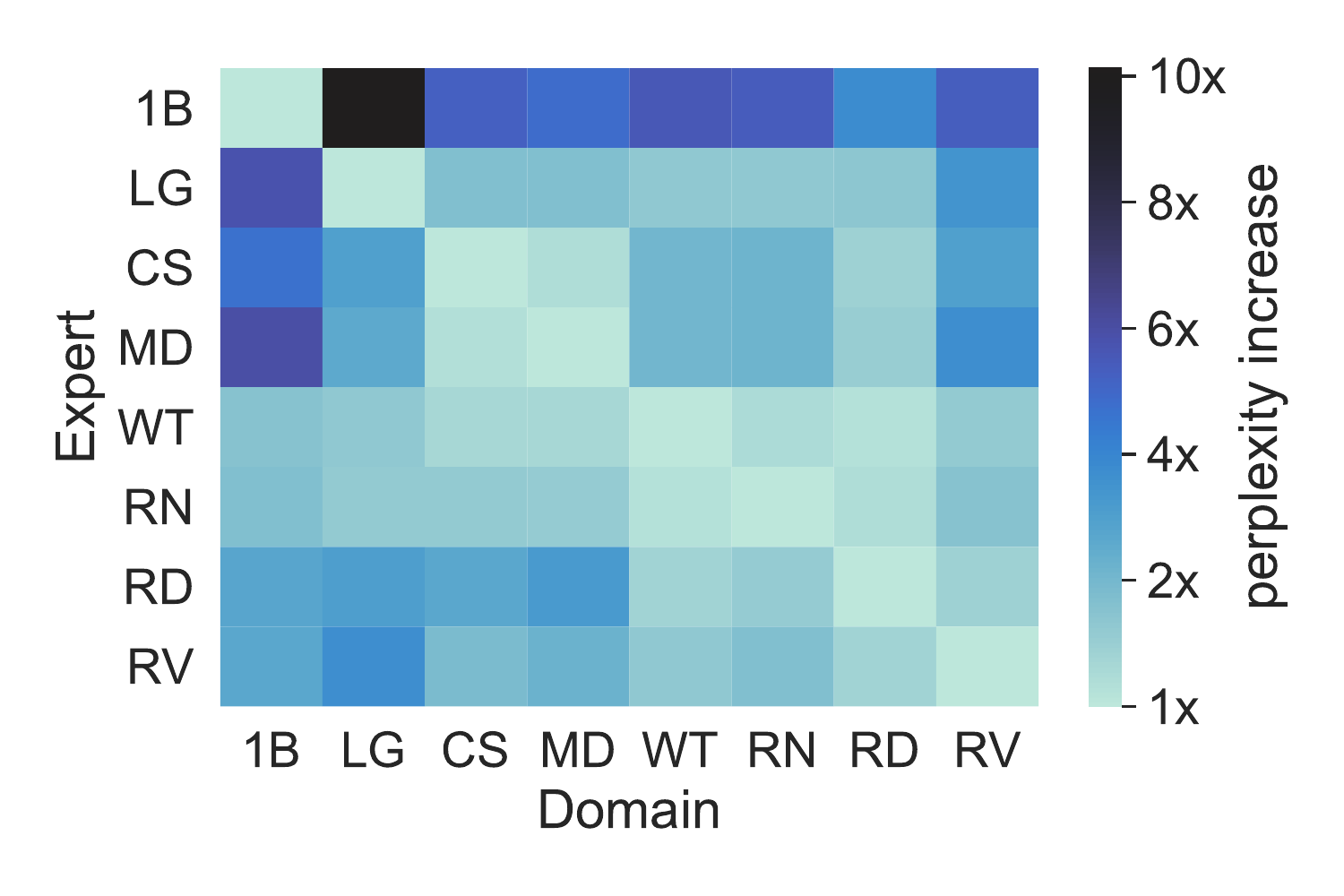}
     \caption{Domain experts in \demix specialize to their domain. We compute the above heatmap with a \demix LM with 1.3B parameters per GPU. Each cell of the heatmap is a ratio between an expert's test perplexity on a domain to that of the expert trained on that domain. The diagonal indicates that each expert has the best performance on its assigned domain. While some experts (e.g., \oneb, \med) do not transfer well to most domains in the training corpus, \webtext and \realnews experts transfer much better, confirming their heterogeneity. Key:  LG $\rightarrow$ \legal, MD $\rightarrow$ Med, WT $\rightarrow$ \webtext, RN $\rightarrow$ \realnews, RD $\rightarrow$ \reddit, RV $\rightarrow$ \reviews.  }
    \label{fig:specialization}
\end{figure}

To provide further evidence for this explanation, we measure the hetereogeneity of domains in the multi-domain corpus, according to a \demix LM. 
We plot a matrix of the perplexity changes across all domain experts in Figure \ref{fig:specialization}, comparing all experts against the expert explicitly trained for each domain. As the perplexity change tends lower, the corresponding expert has higher affinity to the target domain.




First, we observe that domain experts have the highest affinity to their assigned domain, indicating that they do specialize. We also observe that some experts, e.g., \webtext, \realnews, and \reddit, have relatively high affinities to many domains, suggesting that these domains are hetereogeneous. Separately we observe that an expert's affinity to a domain correlates positively with bigram overlap between the expert domain and target domain ($r$=0.40, $t$=3.45, $p$=0.001). This further suggests that similar domains have more closely aligned domain experts.

These findings suggest that a discrete notion of domain, while usually helpful on average (in our artificially constructed population of eight training domains), is too rigid. In the next section, we introduce new ways of softening Equation~\ref{eq:hard} into a mixture over domain experts, to improve performance on heterogeneous domains.

\section{Mixing Experts at Inference Time}
\label{sec:mixing_experts}

The previous section establishes that incorporating \demixlayers improves LM performance on test data from \emph{known} training domains.  At inference time, the domain label was revealed to the model and used to select an expert within each \demix layer.  In practice, however, text may not come with a domain label, may straddle multiple domains, or may not belong to any of the domains constructed at training time; the provenance of the data may even be unknown. 

In these cases, rather than a hard choice among experts (Equation \ref{eq:hard}), we propose to treat $g_1,\ldots,g_n$ as mixture coefficients, transforming the domain membership of an input text into a matter of probabilistic belief.  Unlike previously proposed mixture-of-experts formulations \citep{shazeer2017outrageously, lepikhin2020gshard}, this approach introduces no new parameters and the weights are computed only at test time.\footnote{We choose to explore inference-time mechanisms instead of training mechanisms to mix experts because 1) we want to avoid substantially increasing training costs, i.e., GPU communication between domain experts and 2) we want to maintain the modularity of experts. Exploring mechanisms for training expert  mixtures while satisfying these desiderata is a rich area for future work.}

To analyze inference-time behavior in mixed or unknown domain scenarios, we turn to the corpus of novel domains in the multi-domain corpus (Table \ref{tab:datasets}). As mentioned in \S\ref{sec:dataset}, these domains have fuzzier boundaries, compared to the training domains.

\begin{figure}[t]
    \centering
    \hspace*{-1cm}
    \includegraphics[scale=0.30]{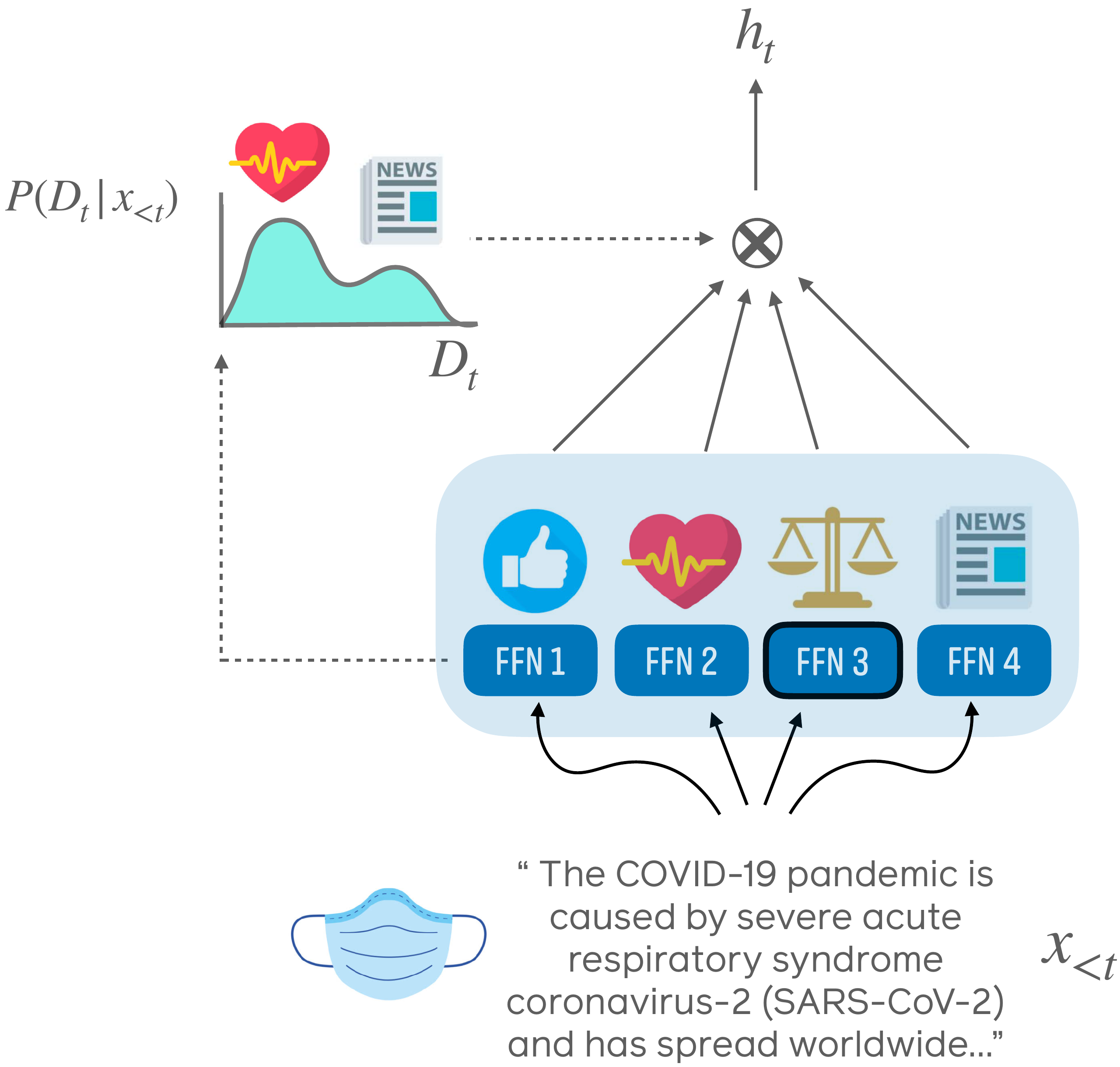}
    \caption{Illustration of inference with domain expert mixing. For a given input text $\boldsymbol{x}_{<t}$ from \cord, we estimate a posterior domain probabilities $p(D_t \mid \boldsymbol{x}_{<t})$, informed by a prior that is either iteratively updated during inference, or is precomputed and cached on held-out data. In this example, the model assigns highest domain probabilities to the medical and news domains. We use these probabilities in a weighted mixture of expert outputs to compute the hidden representation $\mathbf{h}_t$. }
    \label{fig:mixing_experts}
\end{figure}

\subsection{Dynamically Estimating Domain Membership}
\label{sec:bayesian_algo}

Consider the probabilistic view of language modeling, where we estimate $p(X_t \mid \boldsymbol{x}_{<t})$. 
We introduce a domain variable, $D_t$, alongside each word.  We assume that this hidden variable depends on the history, $\boldsymbol{x}_{<t}$, so that:
\begin{align}\small
    p(X_t \mid \boldsymbol{x}_{<t}) & \small{= \sum_{j=1}^n p(X_t \mid \boldsymbol{x}_{<t}, D_t = j) \cdot \underbrace{p(D_t = j \mid \boldsymbol{x}_t)}_{g_j}}
\end{align}
This model is reminiscent of class-based $n$-gram LMs \citep{10.5555/176313.176316} and their derivatives \citep{saul-pereira-1997-aggregate}.

We have already designed the \demix LM to condition on a domain label, giving a form for $p(X_t \mid \boldsymbol{x}_{<t}, D_t = j)$.  The modification is to treat $g_1,\ldots,g_n$ as a posterior probability over domains, calculated at each timestep, given the history so far.

To do this, we apply Bayes' rule:
\begin{flalign}
    \small{p(D_t = j \mid \boldsymbol{x}_t)} &\small{= \frac{p(\boldsymbol{x}_{<t} \mid D_t = j) \cdot p(D_t = j)}{p(\boldsymbol{x}_{<t})}} \\
    &\small{= \frac{p(\boldsymbol{x}_{<t} \mid D_t = j) \cdot p(D_t = j)}{\sum_{j'=1}^n p(\boldsymbol{x}_{<t} \mid D_t = j') \cdot p(D_t = j')}}
\end{flalign}

The conditional probabilities of word sequences given a domain label, as noted above, are already defined by the \demix LM.  For the prior over domain labels, we consider three alternatives:

\begin{figure}[t]
    \centering
    \hspace*{-0.3cm}
    \includegraphics[scale=0.55]{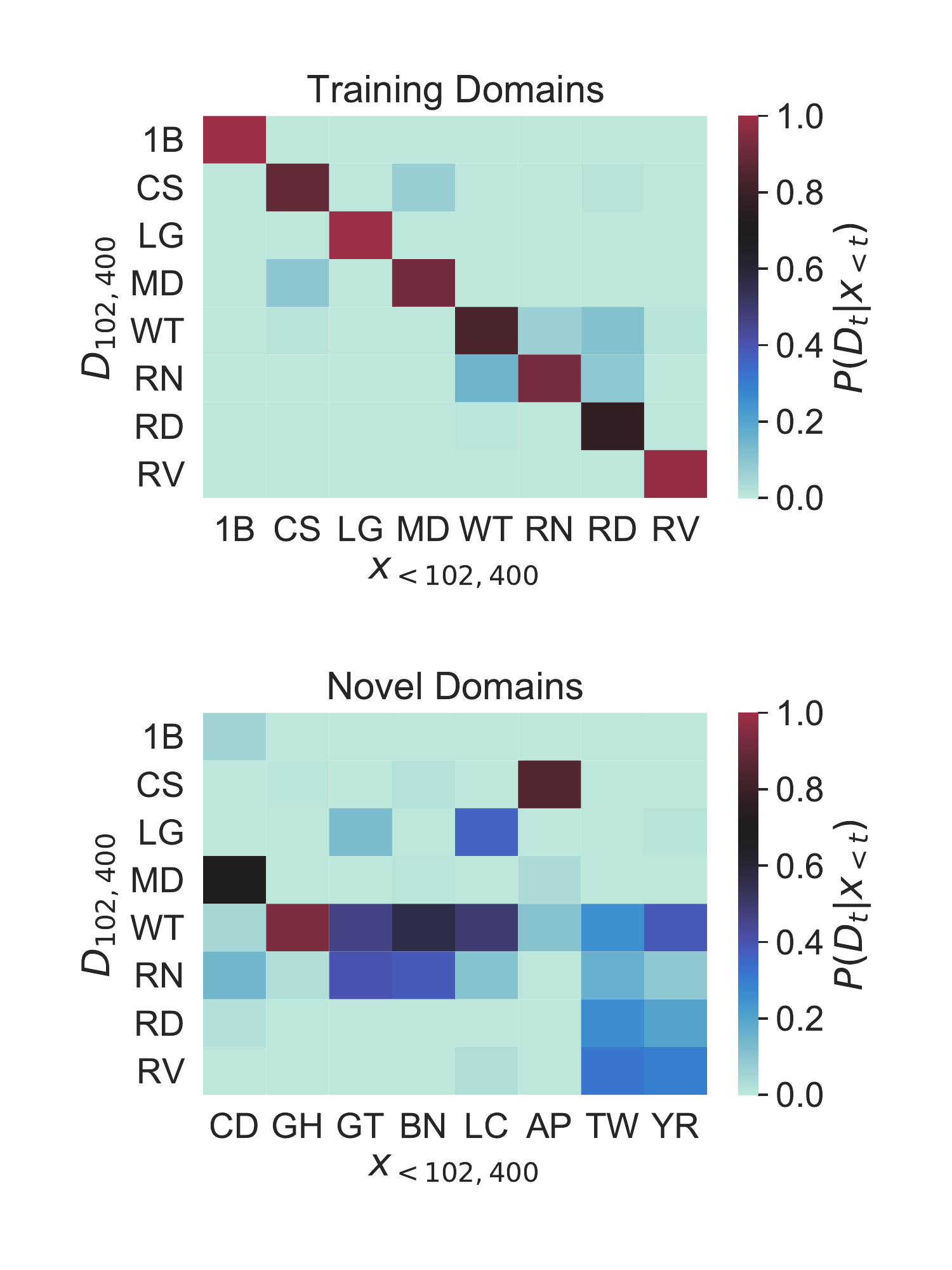}
    \caption{Estimates of posteriors $p(D_t \mid \boldsymbol{x}_{<t})$ with a \demix LM with 1.3B parameters per GPU, after 100 sequences (i.e., 102,400 tokens) of data in training domains (top heatmap) and new domains (bottom heatmap). Key: LG $\rightarrow$ \legal, MD $\rightarrow$ Med, WT $\rightarrow$ \webtext, RN $\rightarrow$ \realnews, RD $\rightarrow$ \reddit, RV $\rightarrow$ \reviews, CD $\rightarrow$ \cord, GH $\rightarrow$ \github, GT $\rightarrow$ \gutenberg, BN $\rightarrow$ \breakingnews, LC $\rightarrow$ \contracts, AP $\rightarrow$ \aclpapers, TW $\rightarrow$ \tweets, YR $\rightarrow$ \yelp. }
    \label{fig:domain_probs_ind_ood}
\end{figure}

\paragraph{Uniform} Fix the prior to be uniform across the known domains.
    
\paragraph{Updating} Set the prior at timestep $t$ to be an exponentially-weighted moving average of the posteriors from previous timesteps: 
    \begin{align}
        p(D_t = j) \propto \sum_{t' =1}^{t-1} \lambda^{t-t'} \cdot p(D_{t'} = j \mid \boldsymbol{x}_{t'})
    \end{align}

 During evaluation, this moving average is calculated over the posterior at the end of each sequence block. The decay factor avoids putting too much weight on calculations made early in the dataset, when posterior calculations are noisier (Appendix \S\ref{sec:domain_posterior_appendix}). We performed a small grid search over \{0.1, 0.3, 0.5, 1.0\} to set the value $\lambda$, and found that 0.3 worked well for most settings.

    
\paragraph{Cached} If, prior to testing, some data from the test distribution is available, we calculate the posterior over domain labels from that data, and fix the prior to that estimate.  Under this setting, we use 100 sequences (i.e., 102,400 tokens) from the development set to estimate the prior, which we found to result in stable posterior probabilities (see Appendix \S\ref{sec:domain_posterior_appendix} for more details).

We display an illustration of the mixture technique in Figure \ref{fig:mixing_experts}.

\subsection{Visualizing Domain Membership}
\label{sec:domain_posteriors}


In Figure \ref{fig:domain_probs_ind_ood}, we plot the posteriors, calculated using the updating method above after 100 sequences of development data, each from training and novel domains. This evaluation is carried out using the \demix LM with 1.3B parameters per GPU from \S\ref{sec:basic_experiment}, with no modifications.

For known domains (top heatmap of Figure~\ref{fig:domain_probs_ind_ood}), the correct label has the highest posterior, but these datasets do not appear to be as distinct or mutually exclusive as we assume. For example,  Reddit data is estimated to be around 80\% \reddit, 11\% \webtext, and 8\% \realnews. More variation in the estimates is expected and observed for the new domains (bottom heatmap of Figure~\ref{fig:domain_probs_ind_ood}).  While \aclpapers is mostly associated with the \cs domain, and \breakingnews mostly with the \webtext and \realnews domains, \cord is spread across \med, \realnews, and \oneb; \yelp across \reviews, \webtext, and \reddit. The alignment of multiple domains like \github and \contracts primarily to \webtext suggests the benefit of including a relatively heterogeneous domain in training.














\begin{table}[t]\small
\centering
\begin{tabular}{rcccc}
\toprule
& \multicolumn{4}{c}{\bf Parameters per GPU}\\
& 125M & 350M & 760M & 1.3B \\
\cmidrule{2-5}
\bf \dense   & 25.9 & 21.4 & 18.4 & 17.8 \\
\bf \dense (B)  & 25.3  & 19.6 &  18.3 & 17.1\\
\bf \domaintoken &  24.8 & 20.4 & 18.4 & 18.0\\ 
\cmidrule{2-5}
\bf \demix (naive) & 28.8 & 23.8 & 21.8 & 21.1 \\ 
\bf \demix (average) & 27.2 & 22.4 & 21.5 & 20.1 \\ 
\bf \demix (uniform) & 24.5 & 20.5 & 19.6 & 18.7 \\ 
\bf \demix (updating) & 21.9 & 18.7 & 17.6 & 17.1\\ 
\bf \demix (cached) & \bf 21.4 & \bf 18.3 & \bf 17.4 & \bf 17.0 \\ 
\bottomrule
\end{tabular}

\caption{Average perplexity on domains unseen during training.  Mixing domain experts with a prior estimated using a small amount of data in the target domain outperforms all other baselines.}
\label{tab:mixing_results}
\end{table}

\subsection{Experimental Setup}

We experiment with the corpus of novel domains (Table \ref{tab:datasets}) to test out-of-distribution performance. We evaluate the three mixture treatments of \demix layers (i.e., uniform, updating, and cached priors) against five baselines.  Note that no new models are trained for this experiment beyond those used in \S\ref{sec:basic_experiment}.

\paragraph {\dense and \dense (Balanced)} These are the basic baselines trained as in \S\ref{sec:basic_experiment}; there is no explicit reasoning about domain.

\paragraph {\domaintoken} Here test data is evaluated using each domain label token, and we choose the lowest among these perplexity values per test set.

\paragraph{\demix (naive)} Similar to \domaintoken, we evaluate the data separately with each of the eight experts, and report the lowest among these perplexity values per test set.

\paragraph{\demix (average)} At every timestep, we take a simple average of the eight experts' predictions.

\subsection{Results}
\label{sec:mixing_results}

\paragraph{Novel Domain Performance}
Results averaged across the eight novel domains are summarized in Table~\ref{tab:mixing_results}. Ensembling \demix experts outperforms \dense baselines and using experts individually (i.e., the ``naive'' baseline), and caching a prior prior to evaluation results in the best average performance. While \domaintoken is competitive with naively using \demixlayers in-domain (Table \ref{tab:base_results}), it consistently underperforms \demix with a weighted mixture on the novel domains. We observe that ensembling \demix experts with a cached prior allows smaller models to match or outperform much larger \dense models. We also find that weighted ensembling outperforms simple averaging, confirming the importance of sparsity in the expert mixture. 

Examining per-domain performance (Appendix \S\ref{sec:per_domain_results}), we find that \demix LMs with a cached prior either outperform \dense baselines or closely match them. The largest improvement against \dense baselines comes from the \tweets domain, which are on average 67\% better across all model sizes. This domain is heterogeneous according to the \demix model (Figure \ref{fig:domain_probs_ind_ood}), confirming the importance of mixing experts for heterogeneous domains. These results demonstrate that conditioning the LM on domains during training need not come at a large cost to generalization to new domains, and in many cases can provide large boosts in performance over \dense baselines.



\paragraph{In-Domain Performance}
\label{sec:improving_indomain}

We can also apply the expert mixture variant of inference (using a cached prior) to the training domains.  We find that doing so is beneficial; see the last line of Table~\ref{tab:base_results}. 

We see improvements in performance across all domains for every scale, though the largest improvements seem to come from hetereogeneous domains (across all model sizes, \reddit improves on average 10.7\%, \webtext 2.4\%, \realnews 1.9\%), again confirming that our intuition that domain metadata may not perfectly align with the most effective domain boundaries. 





\begin{figure}[t]
    \centering
    \hspace*{-0.5cm}
    \includegraphics[scale=0.30]{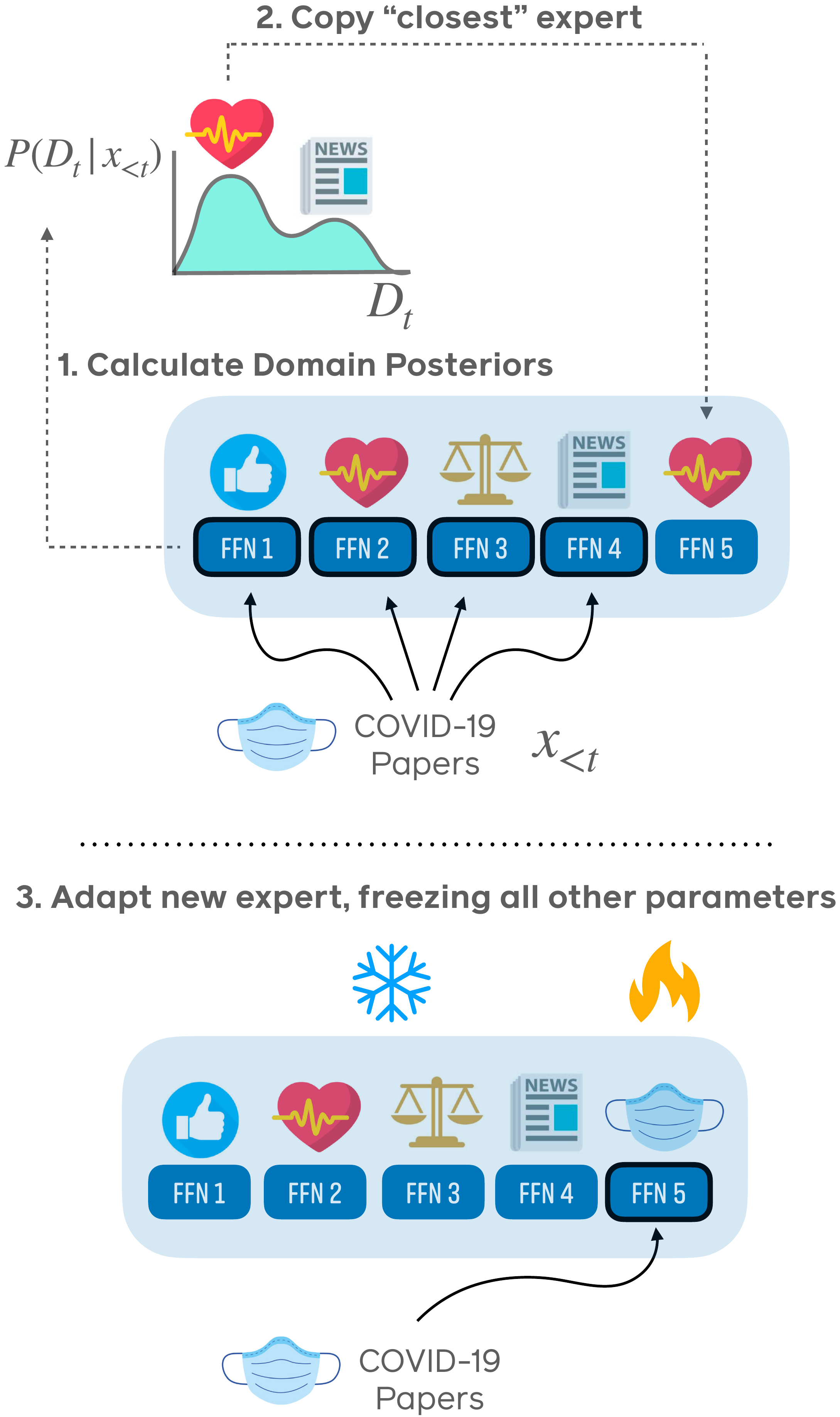}
    \caption{Illustration of \demixdapt. First, we estimate domain posteriors on a held out sample of the target domain (in this case, \cord). We then initialize a new expert with the parameters of the most probable expert under the domain posterior distribution. Finally, we adapt the parameters of the newly initialized expert to the target domain, keeping all other parameters in the LM frozen.} 
    \label{fig:dapt}
\end{figure}

\section{Adaptive Pretraining with New Experts}
\label{sec:adding_experts}



Domain-adaptive, continued pretraining\footnote{This approach typically precedes supervised fine-tuning on task data, hence \emph{pre}training.} of a language model (\dapt) is a way to use unannotated, in-domain text to improve task performance \citep{gururangan-etal-2020-dont}.
However, for a large model, \dapt with \dense training (which we refer to as \densedapt) is expensive and may not be feasible on some computational budgets. Furthermore, \densedapt may result in forgetting what was learned during earlier training phases, limiting reusability.


The modular approach of \demix LMs allows the model to avoid forgetting training domains and adapt cheaply:  we can train a new expert and add it to the \demix layers of the network without updating the other experts or the shared parameters. Because the original model is not changed, forgetting is impossible. We refer to this method of adaptation as \demixdapt.\footnote{Our proposed technique is reminiscent of \emph{Progressive Neural Networks} \citep{rusu2016progressive}.}

We display an illustration of \demixdapt in Figure \ref{fig:dapt}. We instantiate a new expert in each \demix feedforward layer, initialize it with the parameters of the pretrained expert nearest to the new domain. We use the posterior calculations from \S\ref{sec:mixing_experts} on a held-out sample to choose the most probable expert.  We then train the added expert on target data, updating only the new expert parameters. For inference, we use the weighted mixture of domain experts with a cached prior (\S\ref{sec:mixing_experts}). 


\subsection{Experimental Setup}

We compare \demixdapt to \densedapt on all novel domains. We report final test-set perplexity after adapting to each domain for 1 hour with 8 NVIDIA V100 32GB GPUs, tracking validation perplexity every 10 minutes for early stopping. We adapt to each novel domain  with the same hyperparameters as the original phase of training (\S\ref{sec:basic_experiment}), except for a 10x smaller learning rate.

\begin{figure}[t]
    \centering
    \hspace*{-0.5cm}
    \includegraphics[scale=0.5]{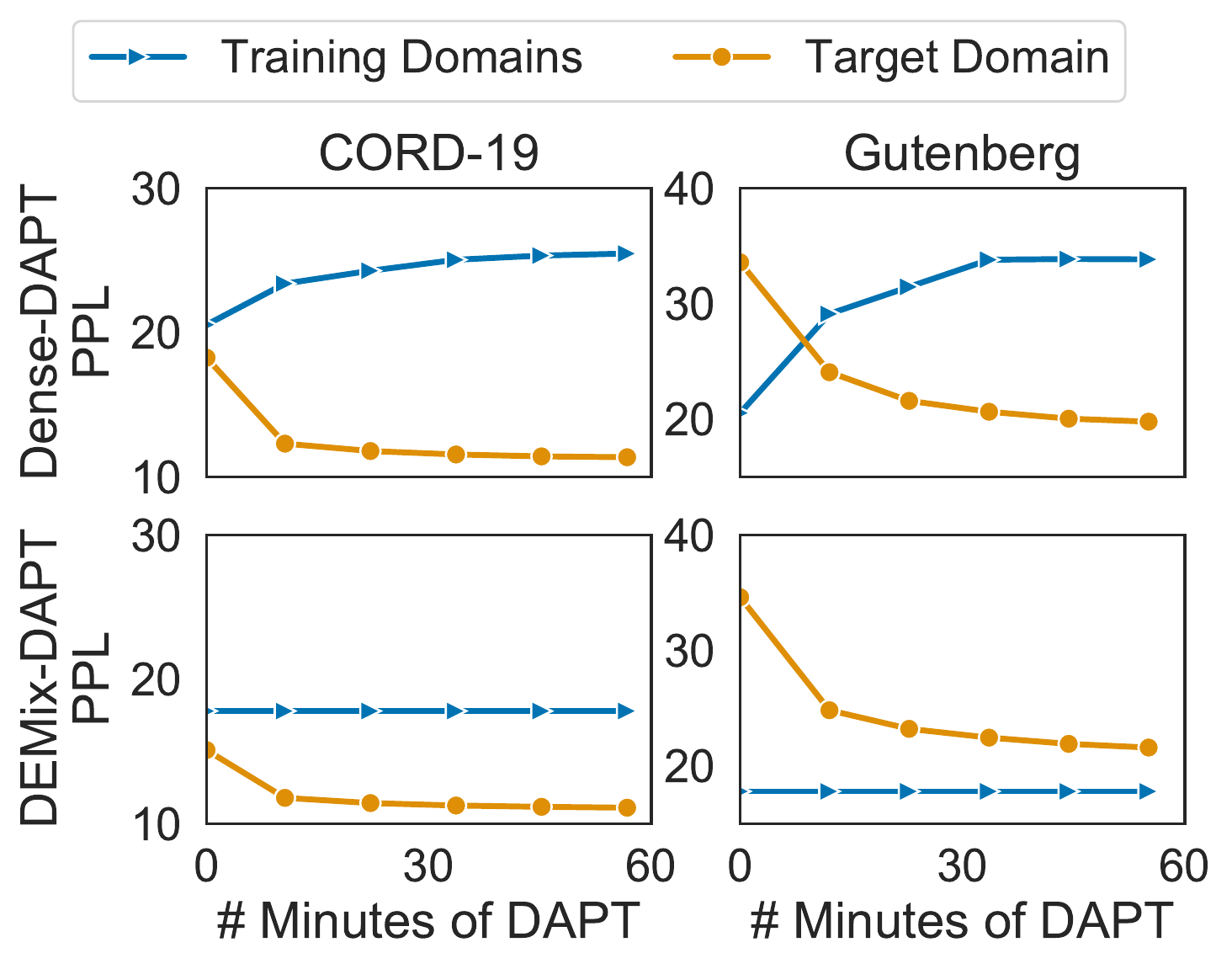}
    \caption{Adapting LMs with 125M parameters per GPU to \cord or \gutenberg.  Top row:  when performing \densedapt on a new domain (\target), average perplexity on all pretraining domains degrades.  Bottom row:  \demixdapt avoids that degradation while achieving close (in the case of \gutenberg) or better (in the case of \cord) performance. The new \cord expert was initialized with the \med expert, and the new \gutenberg expert was initialized with a \webtext expert.} 
    \label{fig:dapt_single}
\end{figure}

\subsection{Results}

\paragraph{Adding one expert} We display examples of \demixdapt and \densedapt on a single additional domain in Figure \ref{fig:dapt_single}. We observe that while \densedapt reduces perplexity on the novel domain, its performance on the training domains progressively worsens, displaying the  forgetting effect (we show similar results in larger models in Appendix \S\ref{sec:dapt_changes}). In contrast, \demixdapt  reduces perplexity on the novel domain \emph{without}  forgetting. 

We generally observe that \demixdapt  outperforms \densedapt for some domains (e.g., \cord and \aclpapers), while it closely approaches \densedapt for others (e.g., \gutenberg; Appendix \S\ref{sec:per_domain_results}). Overall, the parameters for the additional expert comprise about 10\%  of the total parameters in the \demix model, and \densedapt involves updating all the parameters of the model towards in the target domain, so we would expect that \densedapt outperforms \demixdapt in some cases. The strong performance of \demixdapt on domains like \cord and \aclpapers suggests that \demixdapt is especially helpful when the target domain strongly aligns with one of the experts (Figure \ref{fig:domain_probs_ind_ood}). 














\begin{table}[t] \small
\centering
\begin{tabular}{p{1.2cm}ccccc}
\toprule
& &  \multicolumn{4}{c}{\bf Parameters per GPU}\\
\bf Domains & \bf \# Experts & 125M & 350M & 760M & 1.3B \\
\cmidrule{3-6}
\bf \multirow{2}{\linewidth}{\textsc{Training}} & 8 & 17.8 & 14.7 &  13.9 & \bf 13.4 \\ 
& 16 &  \bf 17.7 & \bf 14.6 & \bf 13.7 & \bf 13.4  \\
\cmidrule{3-6}
\bf \multirow{2}{\linewidth}{\textsc{Novel}} & 8 & 21.4 & 18.3 & 17.4 & 17.0 \\
&  16 & \bf 16.0 & \bf 14.0 & \bf 13.5 & \bf 12.5\\



\bottomrule
\end{tabular}

\caption{Average perplexity in training and novel domains before and after adding 8 experts adapted to the novel domains (via \demixdapt). Adding experts reduces perplexity on all domains, even those previously seen.} 
\label{tab:sixteen_experts}
\end{table}

\paragraph{Adding eight experts} With expert mixing (\S\ref{sec:mixing_experts}), newly added experts can be combined with existing ones in the model at test time. To more thoroughly understand the effect of adding more experts to the system, we add all experts adapted to novel domains to the \demix model from \S\ref{sec:basic_experiment}. We display the performance  of a \demix LM with 16 experts (8 experts trained on training domains, 8 additional experts adapted to novel domains) in  Table \ref{tab:sixteen_experts}. We generally observe that \demixdapt reduces perplexity on all domains for all model sizes, again without forgetting.




Adding the eight additional experts in fact reduces perplexity on \emph{previously seen} domains. For example, across all model sizes, on average, we see an 2.4\% reduction on \med, 1.8\% reduction  on \realnews, and 2\% reduction on \reddit (Appendix \S\ref{sec:per_domain_results}). These improvements are small, which is expected given that we only performed \demixdapt for at most one hour with eight GPUs. Even so, these results suggest that \demix layers can enable the LM to incorporate knowledge from novel domains to improve its performance on previously seen domains. 

\section{Language Models with Removable Parts}
\label{sec:removing_experts}

Current LM pretraining datasets are rife with undesirable content, from hatespeech to extremism \citep{gehman-etal-2020-realtoxicityprompts, 10.1145/3442188.3445922}. Another consequence of \dense training is that it is difficult to restrict the model's access to these problematic domains after training, as might be desirable for many user-facing tasks \citep{xu2020recipes, dinan2021anticipating}.


\demixlayers offer new capabilities for lightweight control over the domains in the training data that LMs use to make predictions at inference time. In particular, since \demixlayer experts specialize to their domain (Figure \ref{fig:specialization}), experts that are assigned to domains that are \emph{unwanted} at test-time can be simply disabled and unused.

A key question is whether disabling an expert can simulate a model that has not been exposed to that domain, which we study in this section. However, since the self-attention and input embedding parameters in the \demix LM are shared across domains, removing an expert offers no guarantee of having fully forgotten content from the removed domain. Establishing such bounds is an important avenue for future work.

\begin{table}[t]\small
\centering
\begin{tabular}{lccc}
\toprule
&  \multicolumn{3}{c}{ \bf 125M Parameters per GPU} \\

\cmidrule{2-4}

\textbf{Domain} & \bf \plusexpert  & \bf \minusexpert & \bf \minusdomain \\ 
\midrule 

\oneb

& 13.7 &	25.5 & \bf 30.4  \\ 

\cs 
& 15.7 &	22.4 & \bf 25.4  \\ 

\legal 
& 8.9	& 20.9 &  \bf 22.7    \\ 

\med  
& 12.4 &	 18.6 & \bf 21.9   \\ 

\webtext 
&20.9 &	\bf 27.3 & 25.4   \\

\realnews 
& 18.9 & \bf	26.7 & 25.0  \\ 

\reddit 
& 34.4 &	47.8 & \bf 51.3  \\ 

\reviews 
& 20.5 &	39.0 & \bf 43.0  \\ 

\midrule
\bf Average & 18.2 &	28.5 & \bf 30.6   \\

\bottomrule
\end{tabular}
\caption{In a 125M parameter model, removing a domain expert  (\minusexpert) results in perplexity degradation on the corresponding domain, approaching the performance of an LM that has not been exposed to that domain (\minusdomain). Here we bold the \emph{worst} performing model for each domain, i.e. the one that gets the \emph{highest} perplexity.}
\label{tab:removable_parts_results}
\end{table}

\subsection{Experimental Setup}

To evaluate whether we can simulate models that have not been exposed to a particular domain, we compare three settings:

\paragraph{\plusexpert} A \demix LM with all experts active.

\paragraph{\minusexpert} A \demix LM with a domain expert deactivated.

\paragraph{\minusdomain} A \demix LM retrained from scratch without a particular domain. We replace the removed domain with \gutenberg.\footnote{Our cluster requires that jobs are allocated with eight GPUs, necessitating eight experts --- hence the substitution.}

We evaluate expert removal (\plusexpert and \minusexpert) with the \demix LM with 125M parameters per GPU from \S\ref{sec:basic_experiment}, with no modifications.  For all baselines,we evaluate use expert mixing with a cached prior (\S\ref{sec:mixing_experts}).

\subsection{Results}

Removing a domain expert harms model performance on the associated domain, in most cases approaching the performance of a model that has not been exposed to data from that domain (Table \ref{tab:removable_parts_results}). In some cases (e.g., \webtext and \realnews), \minusexpert even underperforms \minusdomain.  This leads us to conjecture that most domain-specific learning happens within the \demix layer, despite the fact that other parts of the model are affected by all training domains.

\section{Related Work}



\paragraph{Incorporating Metadata} Document metadata has been commonly used to improve the quality of topic models \citep{mimno2012topic, ramage2009, zhu2012}, and previous works have used metadata for adapting RNN-based language models \citep{jaech2018lowrank} or learning better document representations \citep{Card_2018}. \citet{Zellers2019DefendingAN} and \citet{keskar2019ctrl} prepend document metadata in the input text (similar to our \domaintoken setting) while training transformer LMs to provide better inference-time control of text generation.

\paragraph{Inference-time Control} \demixlayers provide a simple mechanism for inference-time control of language model behavior. Previously proposed methods for inference-time control are either expensive to use \citep{Dathathri2020PlugAP}, or rely on densely trained models \citep[e.g.,][]{keskar2019ctrl}. \citet{liu2021dexperts} use multiple experts for inference-time text generation control. This method may be applied to \demix layers to steer text generation with experts trained on different domains.

\paragraph{Multilinguality} Related to variation across domains is crosslingual variation.  Past work has suggested that multilingual models benefit from language-specific parameters  \citep{fan2020englishcentric, pfeiffer-etal-2020-mad, chau-etal-2020-parsing}.  Here, we investigate the effect of incorporating \emph{domain}-specific parameters into the LM.  Though the boundaries between languages are (often) more clear than those among domains, \demixlayers draw inspiration from multilingual research, and future work might explore a compositional approach with both language experts and domain experts. 

\paragraph{Continual Learning} \demixdapt is a type of continual learning, in which the model learns incrementally on new data \citep{8438617}. Previously proposed techniques to support continual learning include regularization \citep{kirkpatrick2017overcoming}, meta-learning \citep{Munkhdalai2017MetaN}, episodic memory modules \citep{lopezpaz2017gradient, dautume2019episodic}, and data replay \citep{sun2019lamol}, all of which may be combined with \demixlayers. Model expansion techniques to incorporate new reinforcement learning or visual tasks \citep{rusu2016progressive, Draelos2017NeurogenesisDL} is especially related to \demixdapt. Our results suggest that continual learning in LMs is naturally enabled with modular domain experts; this may be further explored using temporally-relevant domains \citep{lazaridou2021pitfalls}.

\paragraph{LM Adapters}  Also related to \demixdapt is the line of work into adapter modules for pretrained LMs \citep{houlsby2019parameterefficient, pfeiffer-etal-2020-mad}. Similar to the setting in which we add experts for new domains, adapter modules involve freezing the pretrained language model and updating a small number of additional parameters that are appended to certain parts of the network. This study confirms previous findings that only a subset of LM parameters need to be fine-tuned to a target dataset \citep{zaken2021bitfit}. Expert addition may be performed with adapter modules to further improve efficiency. 

\paragraph{Multi-Domain Models} Multi-domain models have been studied extensively in the context of machine translation, first with statistical systems \citep{banerjee-etal-2010-combining, sennrich-etal-2013-multi}, and more recently with neural networks \citep{Pham2021RevisitingMM}. Other works have explored multi-domain settings with smaller models and explicit domain labels, using supervision \citep[e.g.,][]{wright-augenstein-2020-transformer, guo-etal-2018-multi, zeng-etal-2018-multi} or dense training \citep[e.g.,][]{maronikolakis-schutze-2021-multidomain}. Previous studies have shown the importance considering domains when adapting LMs \citep{ramponi-plank-2020-neural, gururangan-etal-2020-dont}. Our study establishes the importance of considering domains when training LMs from scratch.

\section{Conclusion}

We introduce \demix layers for language models, which provide modularity at inference time, addressing limitations of dense training by providing a rapidly adaptable system. \demixlayers experts can be mixed to handle heterogeneous or unseen domains, added to iteratively incorporate new domains, and removed to restrict unwanted domains. 

There are many exciting directions for future work, in addition to those described throughout the paper. They include combining domain and token-level routing, to realize the benefits of modularity while scaling models efficiently. The design of \demix layers assumes access to coarse provenance labels (or other metadata) to identify domains in pretraining data; an alternative option is to use unsupervised learning to discover domains in the corpus, which, in concert with domain metadata, may lead to better \demix expert assignments. Furthermore, in this work, we study \demix layers with a dataset that has a few large domains. In practice, textual domains usually contain many diverse subdomains of varying prevalence.  Training \demix layers on dataset with a long tail of domains may require automatic measures to cluster smaller domains, or hierarchical experts that are specialized to progressively narrower data distributions. 
\vfill\null

\section*{Acknowledgments}

The authors thank members of UWNLP, FAIR, and AI2, specifically Shruti Bhosale, Tim Dettmers, Emily Dinan, Doug Downey, Margaret Li, Myle Ott, Ofir Press, and Swabha Swayamdipta, for helpful comments.  At UW, this work was partially supported by NSF grant 1562364, the Office of Naval Research under MURI grant N00014-18-1-2670, and an Amazon research award.

\bibliography{anthology,custom}
\bibliographystyle{acl_natbib}

\newpage
\appendix

\section{Appendix}
\label{sec:appendix}

\subsection{Collecting Domains}
\label{sec:domain_collection}

For most domains, we use the associated sources, listed in Table \ref{tab:datasets}, without modification. For \tweets, we use the Twitter Academic API. For \gutenberg, we use the scraping tool provided in \url{https://github.com/aparrish/gutenberg-dammit}. For \breakingnews, we identify a list of factually reliable English news sources, using the list curated by \citet{baly:2018:EMNLP2018}. Specifically, we filter on "high" factuality in the data provided in this repository: \url{https://github.com/ramybaly/News-Media-Reliability}. We then use Newspaper3K (\url{https://newspaper.readthedocs.io/en/latest/}) to scrape the latest 1000 articles from each site. After dropping duplicates, we arrive at about 20K articles from 400 news sources. We provide downloading links and general instructions at \url{https://github.com/kernelmachine/demix-data/blob/main/DOWNLOAD_DATA.md}.

\subsection{Dataset Anonymization}
\label{sec:anonymization}

To anonymize certain datasets, we apply a suite of regexes that aim to identify common patterns of user-identifiable data and substitute them  with dummy tokens. We display anonymization regexes and associated dummy tokens in Table \ref{tab:anonymization}.

\begin{table*}[t]
\centering
\small
\begin{tabular}{lllr}
\toprule
& \bf Category & \bf Link to Regex & Dummy Token  \\
\midrule 
& Email & \url{https://regex101.com/r/ZqsF9x/1} & \texttt{<EMAIL>} \\
& DART & \url{https://regex101.com/r/0tQ6EN/1} & \texttt{<DART>} \\
& FB User ID & \url{https://regex101.com/r/GZl5EZ/1} & \texttt{<FB\_USERID>} \\
& Phone Number & \url{https://regex101.com/r/YrDpPD/1} & \texttt{<PHONE\_NUMBER>} \\
& Credit Card Number & \url{https://regex101.com/r/9NTO6W/1} & \texttt{<CREDIT\_CARD\_NUMBER>} \\
& Social Security Number & \url{https://regex101.com/r/V5GPNL/1} & \texttt{<SSN>} \\
& User handles & \url{https://regex101.com/r/vpey04/1} & \texttt{<USER>} \\
\bottomrule
\end{tabular}

\caption{Anonymization schema. We anonymize text using the regexes provided in the above links for the categories listed.}
\label{tab:anonymization}
\end{table*}

\subsection{Calculating TFLOPs/GPU}
\label{sec:tflops}

We use the formula presented in \citet{narayanan2021efficient} to calculate TFLOPs/GPU and TFLOPs/update. The spreadsheet that contains the calculations and formula can be accessed here: \url{https://docs.google.com/spreadsheets/d/1NO-Lz_VqZGF2fpJTFxtXyjhmaoYi6qnz50Xr8W8hgGw/edit?usp=sharing}.

\subsection{Hyperparameter Assignments}
\label{sec:hyperparameters}

We display hyperparameter assignments for LM pretraining in Tables \ref{tab:125M_hps}, \ref{tab:350M_hps},\ref{tab:760M_hps}, and \ref{tab:1B_hps}.
\begin{table*}[t!]
    \centering
    \small

    \begin{tabular}{cc}
      \toprule
      \textbf{Computing Infrastructure} & 32 Volta 32GB GPUs\\ 
      \bottomrule
    \end{tabular}
    
    \vspace{3mm}\begin{tabular}{cc}
        \toprule
        \textbf{Hyperparameter} & \textbf{Assignment}  \\
        \midrule
        architecture & GPT-3 small \\
        \midrule
        tokens per sample & 1024 \\
        \midrule
        batch size & 2 \\
        \midrule
        number of workers & 2 \\
        \midrule
        learning rate & [5e--4, 3e--4, 1e--4] \\
        \midrule
        clip norm & 0.1 \\
        \midrule
        gradient acculumation steps & 8 \\
        \midrule
        number of steps & 300,000 \\
        \midrule
        save interval updates & 6,000 \\
        \midrule
        validation interval & 3,000 \\
        \midrule
        number of warmup steps & 24,000 \\
        \midrule
        learning rate scheduler & polynomial decay \\
        \midrule
        learning rate optimizer & Adam \\
        \midrule
        Adam beta weights & (0.9, 0.95) \\
        \midrule
        Adam epsilon & 10e-8 \\
        \midrule
        weight decay & 0.1 \\
        \bottomrule
    \end{tabular}
    
    \caption{Hyperparameters for pretraining the LM with 125M parameters per GPU.  All hyperparameters are the same for \demix and \dense training.} 
    \label{tab:125M_hps}
\end{table*}

\begin{table*}[t!]
    \centering
    \small

    \begin{tabular}{cc}
      \toprule
      \textbf{Computing Infrastructure} & 64 Volta 32GB GPUs\\ 
      \bottomrule
    \end{tabular}
    
    \vspace{3mm}\begin{tabular}{cc}
        \toprule
        \textbf{Hyperparameter} & \textbf{Assignment}  \\
        \midrule
        architecture & GPT-3 medium \\
        \midrule
        tokens per sample & 1024 \\
        \midrule
        batch size & 2 \\
        \midrule
        number of workers & 2 \\
        \midrule
        learning rate & [5e--4, 3e--4, 1e--4] \\
        \midrule
        clip norm & 0.1 \\
        \midrule
        gradient acculumation steps & 8 \\
        \midrule
        number of steps & 120,000 \\
        \midrule
        save interval updates & 3,000 \\
        \midrule
        validation interval & 2,000 \\
        \midrule
        number of warmup steps & 9,600 \\
        \midrule
        learning rate scheduler & polynomial decay \\
        \midrule
        learning rate optimizer & Adam \\
        \midrule
        Adam beta weights & (0.9, 0.95) \\
        \midrule
        Adam epsilon & 10e-8 \\
        \midrule
        weight decay & 0.1 \\
        \bottomrule
    \end{tabular}
    
    \caption{Hyperparameters for pretraining the LM with 350M parameters per GPU.  All hyperparameters are the same for \demix and \dense training.} 
    \label{tab:350M_hps}
\end{table*}

\begin{table*}[t!]
    \centering
    \small

    \begin{tabular}{cc}
      \toprule
      \textbf{Computing Infrastructure} & 128 Volta 32GB GPUs\\ 
      \bottomrule
    \end{tabular}
    
    \vspace{3mm}\begin{tabular}{cc}
        \toprule
        \textbf{Hyperparameter} & \textbf{Assignment}  \\
        \midrule
        architecture & GPT-3 large \\
        \midrule
        tokens per sample & 1024 \\
        \midrule
        batch size & 2 \\
        \midrule
        number of workers & 2 \\
        \midrule
        learning rate & [5e--4, 3e--4, 1e--4] \\
        \midrule
        clip norm & 0.1 \\
        \midrule
        gradient acculumation steps & 8 \\
        \midrule
        number of steps & 65,000 \\
        \midrule
        save interval updates & 2,000 \\
        \midrule
        validation interval & 1,000 \\
        \midrule
        number of warmup steps & 5,200 \\
        \midrule
        learning rate scheduler & polynomial decay \\
        \midrule
        learning rate optimizer & Adam \\
        \midrule
        Adam beta weights & (0.9, 0.95) \\
        \midrule
        Adam epsilon & 10e-8 \\
        \midrule
        weight decay & 0.1 \\
        \bottomrule
    \end{tabular}
    
    \caption{Hyperparameters for pretraining the LM with 760M parameters per GPU. All hyperparameters are the same for \demix and \dense training.} 
    \label{tab:760M_hps}
\end{table*}

\begin{table*}[t!]
    \centering
    \small

    \begin{tabular}{cc}
      \toprule
      \textbf{Computing Infrastructure} & 128 Volta 32GB GPUs\\ 
      \bottomrule
    \end{tabular}
    
    \vspace{3mm}\begin{tabular}{cc}
        \toprule
        \textbf{Hyperparameter} & \textbf{Assignment}  \\
        \midrule
        architecture & GPT-3 XL \\
        \midrule
        tokens per sample & 1024 \\
        \midrule
        batch size & 2 \\
        \midrule
        number of workers & 2 \\
        \midrule
        learning rate & [5e--4, 3e--4, 1e--4] \\
        \midrule
        clip norm & 0.1 \\
        \midrule
        gradient acculumation steps & 8 \\
        \midrule
        number of steps & 50000 \\
        \midrule
        save interval updates & 2,000 \\
        \midrule
        validation interval & 500 \\
        \midrule
        number of warmup steps & 4000 \\
        \midrule
        learning rate scheduler & polynomial decay \\
        \midrule
        learning rate optimizer & Adam \\
        \midrule
        Adam beta weights & (0.9, 0.95) \\
        \midrule
        Adam epsilon & 10e-8 \\
        \midrule
        weight decay & 0.1 \\
        \bottomrule
    \end{tabular}
    
    \caption{Hyperparameters for pretraining the LM with 1.3B parameters per GPU. All hyperparameters are the same for \demix and \dense training.} 
    \label{tab:1B_hps}
\end{table*}

\subsection{Per-Domain Results}
\label{sec:per_domain_results}
We display per-domain test results in the spreadsheets at the following link: \url{https://docs.google.com/spreadsheets/d/1yNMZGSPAvhTi3JttLamiCULaOIGTJ4QGEOajO3b5kt8/edit?usp=sharing}

\subsection{Domain Posterior Calculations}
\label{sec:domain_posterior_appendix}

\begin{figure*}
    \centering
    \hspace*{-1cm}
    \includegraphics[scale=0.5]{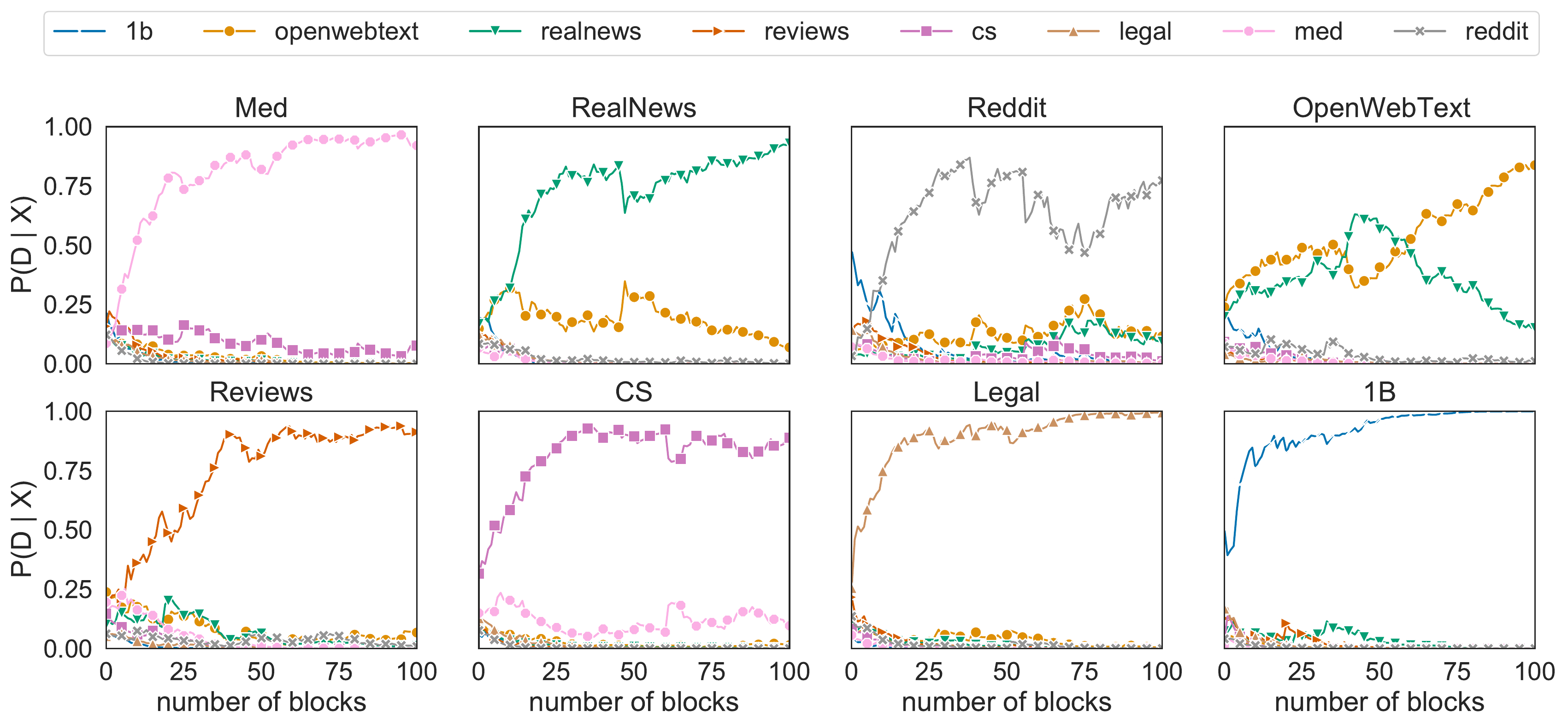}
    \caption{Calculated domain posteriors for 8 training domains. }
    \label{fig:pdx_ind}
\end{figure*}

\begin{figure*}
    \centering
    \hspace*{-1cm}
    \includegraphics[scale=0.5]{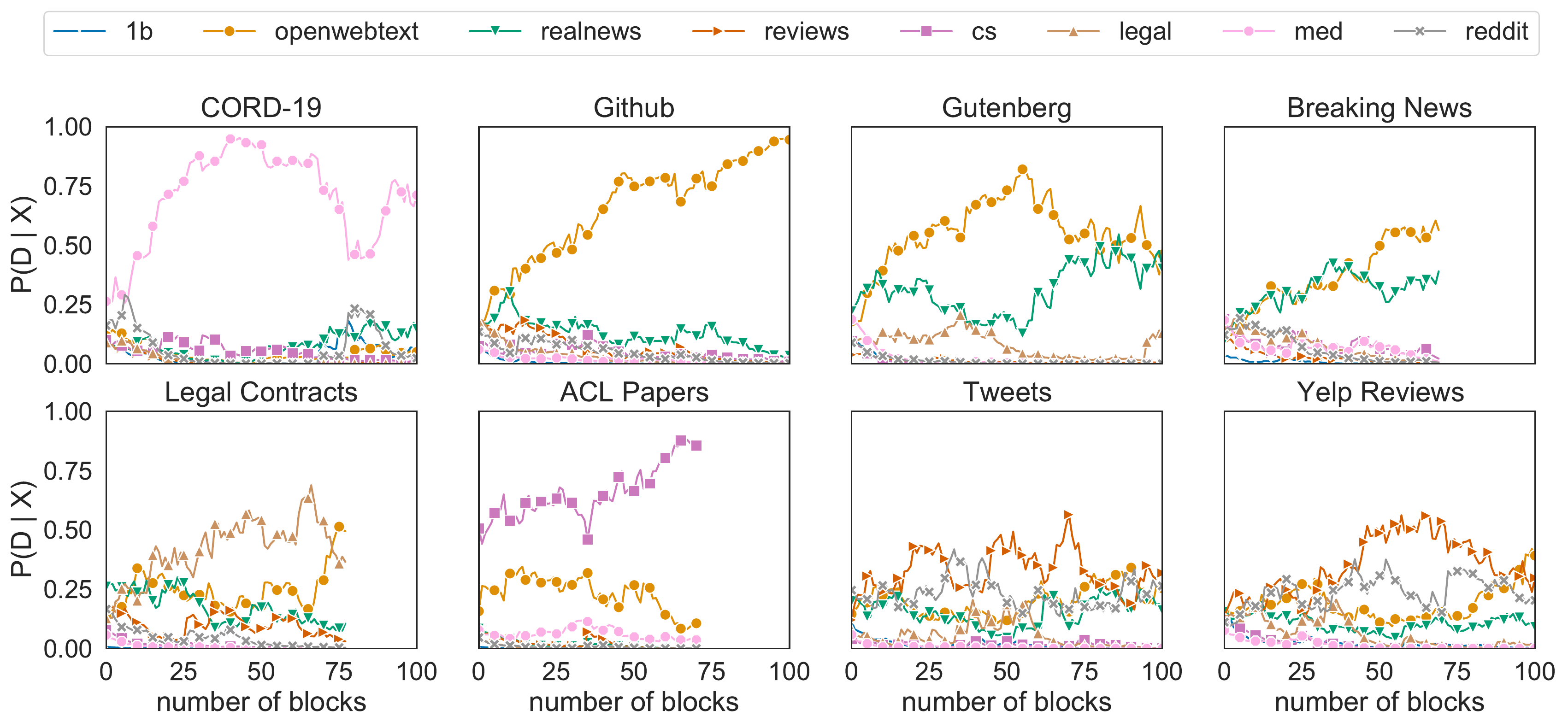}
    \caption{Calculated domain posteriors for 8 novel domains.}
    \label{fig:pdx_ood}
\end{figure*}

We track calculated domain posteriors over blocks of development data in Figure \ref{fig:pdx_ind} (training domains) and Figure \ref{fig:pdx_ood} (novel domains). The calculate domain posteriors are noisier for earlier blocks, stabilizing usually after around 50 blocks. For all experiments, we conservatively use 100 blocks of data to compute the domain posterior, though one may be able to accurately calcuate the domain posterior for some domains with less data.

\begin{table*}[t] \small
\centering
\begin{tabular}{p{1cm}rcccc}
\toprule
& &  \multicolumn{4}{c}{\bf Parameters}\\
& & 125M & 350M & 760M & 1.3B \\
\cmidrule{3-6}
\bf \multirow{2}{0pt}{\densedapt} & T & +70.1\% & +21.4\% &  +16.7\% & +20.6\% \\ 
& N & --55.1\% & --46.6\% & --38.3\% & -44.4\% \\



\bottomrule
\end{tabular}

\caption{Average change in perplexity in training (T) and novel (N) domains after \densedapt.   Negative values indicate better performance relative to the original \dense LM. While average perplexity in the novel domains decreases more for \densedapt, this comes at the cost of a significant deterioration in performance in training domains.} 
\label{tab:dapt_changes}
\end{table*}

\subsection{Perplexity changes after \densedapt}
\label{sec:dapt_changes}

In Table \ref{tab:dapt_changes}, we display the average perplexity change after performing \densedapt on a new domain. We observe that across all model sizes, \densedapt improves  performance in the novel domain, at the cost of a large performance hit in the training domains.

\end{document}